\documentclass[journal]{IEEEtran}

\usepackage{graphicx}
\usepackage{amsmath, amssymb} 
\usepackage{booktabs}        
\usepackage{cite}            
\usepackage{url}             
\usepackage{hyperref}        
\usepackage{algorithm}       
\usepackage{algpseudocode}   
\usepackage{filecontents}    

\title{Generative Inverse Design: From Single-Point Optimization to a Diverse Design Portfolio via Conditional Variational Autoencoders}

\author{Muhammad~Arif~Hakimi~Zamrai,~\IEEEmembership{Student~Member,~IEEE,}
        and~Kamaluddin~Mohd~Yusof,~\IEEEmembership{Member,~IEEE}
\thanks{This work was supported by Universiti Teknologi Malaysia (UTM).}
\thanks{M. A. H. Zamrai and K. M. Yusof are with the Faculty of Electrical Engineering, Universiti Teknologi Malaysia, 81310 Johor Bahru, Johor, Malaysia (e-mail: mahakimi8@graduate.utm.my; kamalmy@utm.my).}}

\begin{document}

\maketitle

\begin{abstract}
Inverse design, which seeks to find optimal parameters for a target output, is a central challenge in engineering. Surrogate-based optimization (SBO) has become a standard approach, yet it is fundamentally structured to converge to a single point solution, thereby limiting design space exploration and ignoring potentially valuable alternative topologies. This paper presents a paradigm shift from single-point optimization to generative inverse design. We introduce a framework based on a Conditional Variational Autoencoder (CVAE) that learns a probabilistic mapping between a system's design parameters and its performance, enabling the generation of a diverse portfolio of high-performing candidates conditioned on a specific performance objective. We apply this methodology to the complex, non-linear problem of minimizing airfoil self-noise, using a high-performing SBO method from a prior benchmark study as a rigorous baseline. The CVAE framework successfully generated 256 novel designs with a 94.1\% validity rate. A subsequent surrogate-based evaluation revealed that 77.2\% of these valid designs achieved superior performance compared to the single optimal design found by the SBO baseline. This work demonstrates that the generative approach not only discovers higher-quality solutions but also provides a rich portfolio of diverse candidates, fundamentally enhancing the engineering design process by enabling multi-criteria decision-making.
\end{abstract}

\begin{IEEEkeywords}
Inverse Design, Generative Models, Conditional Variational Autoencoder (CVAE), Deep Learning, Surrogate-Based Optimization, Bayesian Optimization, Aerodynamics.
\end{IEEEkeywords}

\section{Introduction}
\IEEEPARstart{T}{HE} pursuit of optimal performance is a foundational goal in engineering, from materials science to aerospace. This often manifests as an inverse design problem: given a desired performance metric, what are the system parameters required to achieve it? The primary bottleneck in solving such problems is frequently the high computational cost of the forward evaluation function, $f(\cdot)$, which typically represents a complex physical simulation (e.g., Finite Element Analysis or Computational Fluid Dynamics).

Surrogate-based optimization (SBO) has emerged as a powerful and data-efficient methodology to address this challenge \cite{forrester2008engineering, shahriari2015taking}. SBO approximates the expensive function $f(\cdot)$ with a computationally cheap machine learning model, $\hat{f}(\cdot)$, which then guides an iterative search for the optimum. While effective, SBO frameworks are inherently designed to converge to a single point estimate of the global optimum, $\mathbf{x}^*$. This approach, however, has a critical limitation: the resulting solution, while optimal under the defined objective, may be suboptimal when considering other unmodeled, real-world criteria such as manufacturability, cost, or long-term stability. The engineer is left with a single design and no alternatives.

This paper proposes a fundamental shift from this single-point optimization paradigm to a more holistic, generative approach. We leverage deep generative models to learn the underlying data manifold of the design space and directly synthesize a diverse portfolio of novel, high-performing designs. Specifically, we employ a Conditional Variational Autoencoder (CVAE) \cite{sohn2015learning}, a powerful probabilistic model capable of learning the complex, multi-modal relationship between design parameters and their performance. By conditioning the generative process on a desired performance target, the CVAE can function as a "design synthesizer" producing a multitude of candidates that are predicted to meet a specified objective.

Our primary contribution is the demonstration of this generative framework on a challenging, real-world aerodynamics problem: the minimization of airfoil self-noise \cite{brooks1989airfoil}. We directly compare the portfolio of generated designs against the single best solution found by a top-performing SBO method identified in our previous comprehensive benchmark \cite{hakimi2025benchmark}. We show that the generative approach not only discovers solutions with superior performance but also provides a rich set of diverse alternatives, thereby empowering engineers with the flexibility needed for multi-criteria decision-making.

\section{Theoretical Framework}

\subsection{Problem Formulation}
Let the design space be denoted by $\mathcal{X} \subset \mathbb{R}^d$, where a vector $\mathbf{x} \in \mathcal{X}$ represents a unique set of design parameters. Let $f: \mathcal{X} \to \mathbb{R}$ be the expensive, often black-box, performance evaluation function. The objective of inverse design is to find the optimal parameter set $\mathbf{x}^*$ that minimizes or maximizes the objective function:
\begin{equation}
    \mathbf{x}^* = \arg \min_{\mathbf{x} \in \mathcal{X}} f(\mathbf{x})
\end{equation}
subject to a set of constraints that ensure the physical validity of the design.

\subsection{Conditional Variational Autoencoders (CVAEs)}
Our generative framework is built upon the CVAE, which extends the standard Variational Autoencoder (VAE) \cite{kingma2014auto}. A VAE is a generative model that learns a probabilistic mapping from a high-dimensional data space $\mathcal{X}$ to a lower-dimensional, continuous latent space $\mathcal{Z}$. It consists of an encoder (or recognition model) $q_\phi(\mathbf{z}|\mathbf{x})$ and a decoder (or generative model) $p_\theta(\mathbf{x}|\mathbf{z})$, where $\phi$ and $\theta$ are the parameters of neural networks.

The VAE is trained by maximizing the Evidence Lower Bound (ELBO) on the marginal log-likelihood of the data:
\begin{equation}
    \mathcal{L}(\theta, \phi; \mathbf{x}) = \mathbb{E}_{q_\phi(\mathbf{z}|\mathbf{x})}[\log p_\theta(\mathbf{x}|\mathbf{z})] - \beta D_{KL}(q_\phi(\mathbf{z}|\mathbf{x}) || p(\mathbf{z}))
    \label{eq:elbo}
\end{equation}
The first term in Eq. \ref{eq:elbo} is the reconstruction loss, which encourages the decoder to accurately reconstruct the input data. The second term is the Kullback-Leibler (KL) divergence between the approximate posterior from the encoder and a prior over the latent variables, $p(\mathbf{z})$, which is typically a standard normal distribution $\mathcal{N}(0, \mathbf{I})$. The hyperparameter $\beta$ is introduced to control the weight of the KL divergence term.

A CVAE extends this framework by conditioning both the encoder and decoder on an additional attribute vector $\mathbf{c}$. In our application, $\mathbf{x}$ is the vector of airfoil design parameters and $\mathbf{c}$ is the scalar performance value $y = f(\mathbf{x})$. The model learns the joint probability distribution $p(\mathbf{x}, \mathbf{z} | \mathbf{c})$. The CVAE's objective function becomes:
\begin{equation}
\begin{split}
    \mathcal{L}(\theta, \phi; \mathbf{x}, \mathbf{c}) = \mathbb{E}_{q_\phi(\mathbf{z}|\mathbf{x},\mathbf{c})}[\log p_\theta(\mathbf{x}|\mathbf{z},\mathbf{c})] \\ - \beta D_{KL}(q_\phi(\mathbf{z}|\mathbf{x},\mathbf{c}) || p(\mathbf{z}|\mathbf{c}))
\end{split}
\label{eq:cvae_elbo}
\end{equation}
Assuming the prior $p(\mathbf{z}|\mathbf{c})$ is independent of the condition, it simplifies to $p(\mathbf{z})$. Once trained, the CVAE can be used as a generative model. By fixing the condition to a desired target performance, $\mathbf{c}_{target}$, and sampling a latent vector $\mathbf{z}_{sample}$ from the prior distribution $\mathcal{N}(0, \mathbf{I})$, we can generate a new design $\mathbf{x}_{new}$ via the decoder:
\begin{equation}
    \mathbf{x}_{new} \sim p_\theta(\mathbf{x}|\mathbf{z}_{sample}, \mathbf{c}_{target})
\end{equation}
This allows us to synthesize an entire distribution of designs that are all predicted to achieve the target performance.

\begin{figure*}[!t]
\centering
\includegraphics[width=\textwidth]{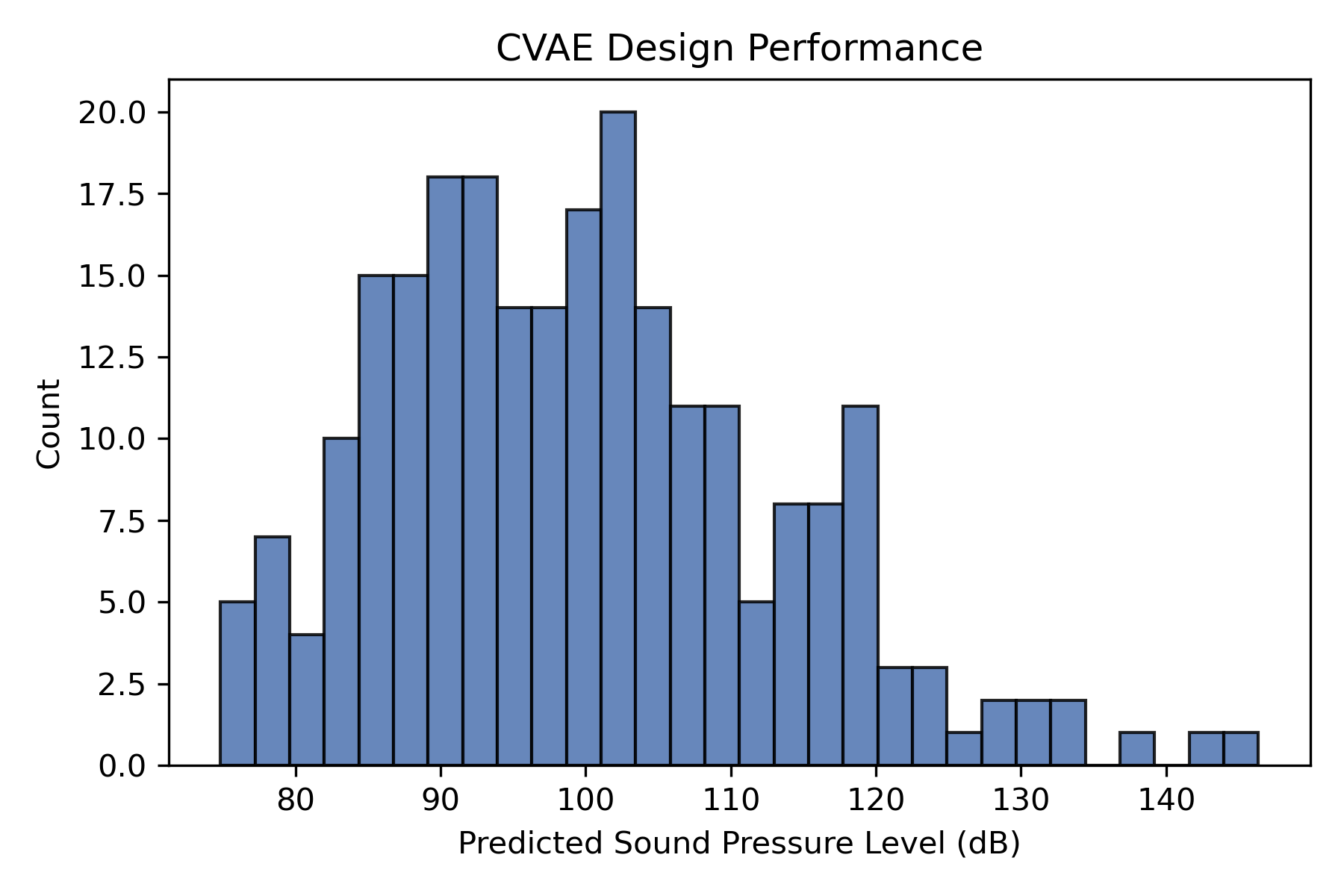}
\caption{Performance distribution of the 241 valid airfoil designs generated by the CVAE, as evaluated by a trained MLP surrogate model. The vertical dashed red line at 108.81 dB indicates the surrogate-predicted performance of the single best design found by the top-performing SBO method from our previous benchmark \cite{hakimi2025benchmark}. A substantial majority of the generated designs (186 of 241, or 77.2\%) fall to the left of this baseline, demonstrating superior (lower noise) performance. The generative approach not only finds better solutions but also provides a diverse portfolio of high-quality alternatives.}
\label{fig:cvae_hist}
\end{figure*}

\section{Experimental Methodology}

\subsection{Benchmark Task: Airfoil Self-Noise}
We selected the NASA airfoil self-noise dataset \cite{brooks1989airfoil} as our benchmark task due to its well-established use and non-linear characteristics. The dataset comprises 1,503 measurements from wind tunnel tests. The design vector $\mathbf{x} \in \mathbb{R}^5$ includes features such as frequency, angle of attack, and chord length. The objective is to minimize the output $y \in \mathbb{R}$, which represents the scaled sound pressure level in decibels (dB). All input features were standardized to have zero mean and unit variance before training.

\subsection{Performance Baseline: SBO}
To provide a rigorous point of comparison, we use the results from our prior, comprehensive SBO benchmark on this same task \cite{hakimi2025benchmark}. In that work, we evaluated five surrogate models under various data and budget constraints. The best-performing method for this task under realistic constraints (a small initial dataset) was an SBO loop using a Linear surrogate model, which discovered a design with a true, validated performance of 103.38 dB. This single design represents the SBO performance baseline for this study.

\subsection{CVAE Implementation and Generative Campaign}
We implemented the CVAE using PyTorch. The architecture for both the encoder and decoder consisted of two fully-connected hidden layers with 128 neurons each and ReLU activation functions. The latent space dimensionality was set to $d_z=8$. The model was trained for 400 epochs using the Adam optimizer with a learning rate of $10^{-3}$ and a batch size of 128. The KL divergence weight was set to $\beta=1.0$.

For the generative campaign, we defined our target condition $\mathbf{c}_{target}$ as the 10th percentile of the sound pressure levels in the training data, which corresponds to a high-performance (low-noise) value of 115.08 dB. We then generated a portfolio of $N=256$ new airfoil designs by sampling from the latent prior and conditioning the decoder on this target.

\subsection{Evaluation Metrics and Procedure}
The generated portfolio was assessed using three quantitative metrics:

\subsubsection{Validity} A generated design $\mathbf{x}_{gen}$ is considered valid if each of its features, after inverse scaling, falls within the physical bounds of the original training data. We define these bounds with a 5\% margin to allow for reasonable extrapolation. Let $\mathbf{x}_{min}$ and $\mathbf{x}_{max}$ be the feature-wise minimum and maximum vectors from the training data, and let $\mathbf{\Delta} = 0.05(\mathbf{x}_{max} - \mathbf{x}_{min})$. A design is valid if for every feature $j$:
\begin{equation}
    x_{gen, j} \in [x_{min, j} - \Delta_j, x_{max, j} + \Delta_j]
\end{equation}

\subsubsection{Diversity} To quantify the novelty of the generated designs, we compute the average pairwise Euclidean distance, a measure of spread in the design space:
\begin{equation}
    D = \frac{1}{\binom{n_v}{2}} \sum_{i=1}^{n_v-1} \sum_{j=i+1}^{n_v} ||\mathbf{x}_i - \mathbf{x}_j||_2
\end{equation}
where $n_v$ is the number of valid designs and the vectors $\mathbf{x}_i$ are the unscaled design parameters.

\subsubsection{Performance} As direct simulation of the generated designs is infeasible, we trained a high-fidelity MLP surrogate model on the entire 1,503-point dataset to serve as a consistent performance oracle. This surrogate allows for a fair, apples-to-apples comparison by providing performance predictions ($\hat{y}$) for both the newly generated designs and the SBO baseline design. The MLP's strong predictive accuracy on this dataset was established in our prior work \cite{hakimi2025benchmark}.

\section{Results and Analysis}
The CVAE-based generative process yielded a large portfolio of novel, valid, and high-performing designs that demonstrably surpassed the SBO baseline.

\subsection{Portfolio Generation and Characteristics}
The generative campaign produced 256 unique airfoil designs. Of these, 241 (a validity rate of 94.1\%) met the physical validity criteria. The remaining 15 invalid designs marginally violated constraints on chord length and velocity.

The portfolio of 241 valid designs exhibited significant diversity, with an average pairwise distance of 2491.6 in the original feature space. This high value confirms that the CVAE did not merely memorize and reproduce training samples, but rather learned a continuous representation of the design space from which it could synthesize genuinely novel candidates.

\subsection{Quantitative Performance Analysis}
The 241 valid designs were evaluated using the pre-trained MLP surrogate. The resulting performance distribution is shown in Fig. \ref{fig:cvae_hist}. The mean predicted sound level for the generated portfolio was 99.74 dB, with a standard deviation of 13.60 dB. The top-performing generated design achieved a predicted sound level of 74.83 dB, significantly lower than any value present in the original dataset.

The central finding of this study is the direct comparison with the SBO baseline. The SBO baseline design (with a true value of 103.38 dB) received a predicted score of 108.81 dB from our MLP surrogate. As shown in Fig. \ref{fig:cvae_hist}, a remarkable **186 out of the 241 valid designs (77.2\%) achieved a predicted score lower than this baseline**. This result provides strong evidence for the superiority of the generative approach.

\section{Discussion}
The results highlight a fundamental advantage of generative inverse design over traditional single-point optimization.

\subsection{Learning the Design Manifold vs. Searching for a Point}
SBO methods operate by sequentially navigating the design space, using a surrogate to propose the next most promising point to evaluate. This process is inherently local and path-dependent, designed to converge to a single optimum. In contrast, the CVAE learns a global, probabilistic representation of the entire high-performance design manifold. It does not search for a single point but rather learns the underlying "rules" that govern what constitutes a good design. This holistic understanding allows it to generate diverse solutions across this manifold in a single, parallelizable step, a far more efficient and comprehensive approach to design space exploration.

\subsection{The Engineering Value of a Design Portfolio}
The practical implication of generating a portfolio of 186 superior designs, rather than finding one, is profound. This portfolio empowers engineers with the flexibility to perform multi-criteria decision-making. For example, from this set of high-performing airfoils, a designer could select the one that is easiest to manufacture, has the most favorable structural properties, or is most robust to variations in operating conditions. The generative approach decouples the primary performance optimization from these other critical, real-world considerations, thereby streamlining and enriching the entire engineering workflow.

\subsection{Limitations and Future Work}
While promising, this study has limitations. The performance evaluation relies on a surrogate model; validation through high-fidelity simulations would be a necessary next step for real-world deployment. Furthermore, the scalability of this CVAE approach to problems with significantly higher dimensionality ($d \gg 10$) or discrete parameter spaces remains an open area for investigation. Future work could explore integrating this generative model within an active learning loop, where the CVAE proposes a batch of diverse, high-performing candidates, which are then evaluated and used to refine the model in subsequent rounds.

\section{Conclusion}
This paper introduced a generative inverse design framework using a Conditional Variational Autoencoder and demonstrated its superiority over traditional surrogate-based optimization on a complex aerodynamics problem. By learning a conditional mapping of the design space, the CVAE was able to generate a diverse portfolio of 241 valid and novel airfoil designs. A rigorous, surrogate-based comparison showed that over 77\% of these designs outperformed the single best solution found by a state-of-the-art SBO method.

The key insight is the shift from finding a single answer to generating a diverse set of high-quality solutions. This work establishes generative models not merely as an alternative, but as a more powerful and practical tool for solving modern engineering design challenges, paving the way for accelerated innovation and more robust, multi-faceted design decisions.

\bibliographystyle{IEEEtran}
\bibliography{references}

\end{document}